\documentclass[letterpaper, 10 pt, conference]{ieeeconf}  % Comment this line out if you need a4paper
\usepackage{mathtools}
\usepackage{import}
\usepackage{times}
\usepackage{fancyhdr,graphicx,amsmath,amssymb}
\usepackage[ruled,vlined]{algorithm2e}
\usepackage{multirow}
\usepackage{colortbl}
\usepackage{rotating}
\usepackage[dvipsnames]{xcolor}
\usepackage{float}

\IEEEoverridecommandlockouts                              % This command is only needed if 
\title{\LARGE \bf
Using 3-D LiDAR Data for Safe Physical Human-Robot Interaction
}

\author{Sarthak Arora$^{1}$, Karthik Subramanian$^{2}$, Odysseus Adamides$^{3}$, and Ferat Sahin$^{4}$% <-this % stops a space
\thanks{This material is based upon work supported by the National Science Foundation under Award No. DGE-2125362. Any opinions, findings, and conclusions or recommendations expressed in this material are those of the author(s) and do not necessarily reflect the views of the National Science Foundation.}% <-this % stops a space
\thanks{$^{1}$Sarthak Arora is with the Department of Electrical Engineering and Micro-electronic, Rochester Institute of Technology, 1 Lomb Memorial Drive, Rochester, NY 14623, USA {\tt\small sa9472@rit.edu}}%
\thanks{$^{2}$Karthik Subramanian is with the Department of Electrical and Micro-electronic Engineering, Rochester Institute of Technology, 1 Lomb Memorial Drive, Rochester, NY 14623, USA {\tt\small kxs8997rit.edu}}%
\thanks{$^{3}$Odysseus Adamides is with the Department of Electrical and Micro-electronic Engineering, Rochester Institute of Technology, 1 Lomb Memorial Drive, Rochester, NY 14623, USA {\tt\small oaa8092@rit.edu}}%
\thanks{$^{4}$Ferat Sahin is with Faculty of Electrical and Microelectronic Engineering, Rochester Institute of Technology, 1 Lomb Memorial Drive, Rochester, NY 14623, USA {\tt\small feseee@rit.edu}}%
}

\begin{document}

\maketitle
\thispagestyle{empty}
\pagestyle{empty}

%%%%%%%%%%%%%%%%%%%%%%%%%%%%%%%%%%%%%%%%%%%%%%%%%%%%%%%%%%%%%%%%%%%%%%%%%%%%%%%%
\begin{abstract}
This paper explores the use of 3D lidar in a physical Human-Robot Interaction (pHRI) scenario. To achieve the aforementioned, experiments were conducted to mimic a modern shop-floor environment. Data was collected from a pool of seventeen participants while performing pre-determined tasks in a shared workspace with the robot. To demonstrate an end-to-end case; a perception pipeline was developed that leverages reflectivity, signal, near-infrared, and point-cloud data from a 3-D lidar. This data is then used to perform safety based control whilst satisfying the speed and separation monitoring (SSM) criteria. In order to support the perception pipeline, a state-of-the-art object detection network was leveraged and fine-tuned by transfer learning. An analysis is provided along with results of the perception and the safety based controller. Additionally, this system is compared with the previous work.
\end{abstract}
%%%%%%%%%%%%%%%%%%%%%%%%%%%%%%%%%%%%%%%%%%%%%%%%%%%%%%%%%%%%%%%%%%%%%%%%%%%%%%%%
\section{Introduction}

Industry 4.0 has significantly increased the integration of point rich perception sensors into industries including manufacturing, supply chain, warehousing, medical fields, and construction \cite{xu_industry_2018}. The integration of these sensors has expanded the automation capabilities of these fields. A key sensor technology integrated across these fields has been lidar. These sensors provide two dimensional and three dimensional information about the environment around them and have been used to detect objects, obstacles, and humans through processes and tasks going on in the workspace around them \cite{mcmanamon_lidar_2019}. With the data provided by lidars, the industry has been able to implement more complex algorithms and autonomous approaches within their fields. This includes the rise of autonomous vehicles, autonomous space vehicle landings, automated guided vehicles  (AGVs), unmanned areal vehicles (UAVs), and collaborative robotics applications \cite{kumar_lidar_2017,wang_pedestrian_2017,munasinghe_covered_2022,rozsa_obstacle_2018,balasa_lidar_2021}. Throughout the past decade, both the algorithms and sensors have continued to see significant innovations. 
\begin{figure}[H]
    \centering
    \includegraphics[width=8.5cm]{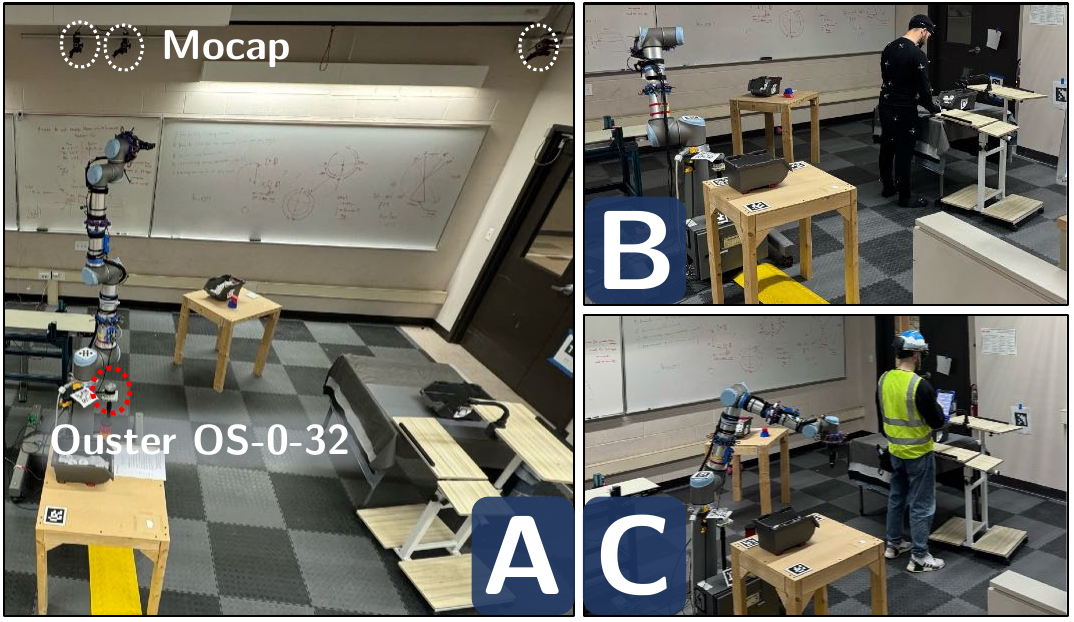}
    \caption{An image showing different stages of the experiment setup used in this work. In image ``A", the layout of the robot workspace is shown along with the exteroceptive sensors used in the setup (encircled in red and white). In image ``B", the test subject is wearing a motion capture body suit for acquiring minimum distance associated with the human and robot. In image ``C", the participant is wearing a reflective vest and reflective hardhat.}
    \label{fig:cover}
\end{figure}

Time-of-Flight (ToF) cameras have begun to increase in depth resolution, it has become easier to calculate depth for stereoscopic cameras, and millimeter wave radar has begun to be used for human tracking applications \cite{tan_triple_2011,adamides_evaluation_2023,lacevic_safe_2023,ubezio_radar_2021}. Though there has been a diversification in perception options, lidar remains to be a commonly used sensing method across industrial and research applications \cite{peters_extrinsic_2020,rashid_local_2020,behera_urban_2022,karypidis_point_2023}. In the past few years, there has been releases of new lidar products lines which bring new innovations to the perception platform. One product line example was released by the lidar manufacturer Ouster. 
% \begin{figure}[H]
%     \centering
%     \includegraphics[width=8.5cm]{cover.pdf}
%     \caption{An image showing different stages of the experiment setup used in this work. In image ``A", the layout of the robot workspace is shown along with the exteroceptive sensors used in the setup (encircled in white). In image ``B", the test subject is wearing motion capture body suit for acquiring minimum distance associated with the human and robot. In image ``C", the participant is wearing a reflective vest and reflective hardhat along with an IR emitting XR device.}
%     \label{fig:cover}
% \end{figure}
 The “OS” series of lidars includes the OS0, OS1, OS2, and OSDome. Along with various viewing angles and data channels, this product line generates four 2D image data modalities formed from the traditional 3D point cloud the lidar generates \cite{shi_object_2022}. The four frames are range, signal, Near-IR (infra-red), and Reflectivity frames. The Range frame provides a per-pixel ToF distance calculation from the sensor origin to the pixel in the Range frame. The Signal frame provides the light return strength per pixel in the frame. The Near-IR frame provides the light return to the sensor per pixel that was not generated by the laser emitter local to the lidar. This frame measurement is similar to a monochrome IR return from a traditional image sensor. Lastly, the reflectivity frame provides the reflectivity strength per pixel. This frame provides key data on the reflectivity of materials and surfaces in the environment. The significance of Ouster including these frames in addition to the traditional point cloud is that it allows existing 2D machine learning algorithms to be directly applied to the 3D lidar data \cite{shi_object_2022}. Hence, by leveraging the data provided by the channels, we aim to make the following contributions:
% Our goal is to evaluate (and ultimately, leverage) the use of the 3D lidar affixed to the robot base (on-robot) in a state-of-the-art industrial shop-floor setup. 
\begin{enumerate}
    \item Develop a lidar based dataset with multiple participants with varying clothing and body shapes in realistic shop-floor conditions. 
    % REJECTED \textit{We hypothesize that varying clothing colors and materials will affect the system performance due to varying surface reflectance levels.}
    \item Demonstrate a successful use of the data collected by training a state-of-the-art object detector with validation and testing.
    \item An application of the perception pipeline to develop a simple speed and separation monitoring safety controller based on prior work in \cite{iso_isots_2022}
\end{enumerate}

%this explains what is already illustrated in the paragrpah above. which way do we want to express this information?
% Range [32 bit unsigned int - only 20 bits used] - range in millimeters, discretized to the
% nearest 3 millimeters.
% Signal Photons [16 bit unsigned int] - signal intensity photons in the signal return measure-
% ment are reported.
% Reflectivity [16 bit unsigned int] - sensor Signal Photons measurements are scaled based
% on measured range and sensor sensitivity at that range, providing an indication of target
% reflectivity. Calibration of this measurement has not currently been rigorously implemented,
% but this will be updated in a future firmware release.
% Near Infrared Photons [16 bit unsigned int] - NIR photons related to natural environmental
% illumination are reported.

\section{Literature Survey}
2D frames of 3D lidar point clouds have been used in a number of research fields. This includes \cite{zhang_ri-lio_2023}, where the reflectivity image was used to correct drone odometry. Additionally, \cite{man_multi-echo_2021} fuses the multiple modalities to increase 3D object detection performance. These alternate frame data formats have also been used in the automotive field to test segmentation of humans, vehicles, and other traffic objects without the use of a traditional CMOS image sensors \cite{tsiourva_lidar_2020}. In these different applications, there are plenty of previous works that illustrate a sufficient approach for feature extraction and data formatting to feed image based classifiers and algorithms.
With the dawn of Industry 5.0, it is imperative for lidar to maintain compatibility with 2D machine vision and machine learning algorithms such that lidars match the performance of other perception systems used across industry \cite{barata_industry_2023}. Industry 4.0 setup the infrastructure of digitally driven and automated processes, Industry 5.0 pushes researchers to look deeper at these processes and their impact on the human individuals who must coexist with this infrastructure. A key research area that will continue to be a focus area in Industry 5.0 is Human Robot Collaboration (HRC). In this field, the pose of the worker, distance from worker to robot, and trajectories of the human and robot in the workspace are vital to increasing the safety and comfort of the worker \cite{kumar_speed_2019,villani_survey_2018,kumar_survey_2021}. Speed and separation monitoring (SSM) is one of the four major collaborative approaches identified in the International Organization for Standardization (ISO) standard ISO/TS 15066:2016 \cite{iso_isots_2022}.
In the field of SSM research, a number of different sensor configurations and modalities are considered including ToF cameras, stereo cameras, mmWave radars, ultrasonic sensors, and lidars \cite{kumar_survey_2021}. Lidar was the primary sensor used in the early years of SSM research \cite{marvel_implementing_2017}. As innovations in computation and perception have progressed, the other perception modalities have seen a rise of use in the field. To track the human in the scene, it is crucial for the image based perception systems used in an SSM setup to feed data to convolutional neural networks (CNNs) \cite{patalas-maliszewska_working_2024}. This localization of the human in the frame enables the computation of minimum distance data needed for an SSM algorithm. With the Ouster OS-0-32, the lidar data can also be used to directly feed CNN based algorithms for human position, and pose tracking.

In this paper, frame based lidar data is directly used to train a YOLOv9 \cite{wang_yolov9_2024} model in contrast to traditional methods which require raw 3D point cloud processing and mapping prior to the input into a neural network. Additionally, the data captured in this work consists of diverse body shapes and clothing material in an industrial environment. Furthermore, the data and model is applied to a simple, generalized SSM algorithm which outputs a safety distance and an operational velocity scaling factor. Lastly, the paper explores the viability of a vertical and horizontal field of view (FoV) lidars for safety based applications. The dataset, and trained model will be provided for other researchers to conduct further studies.

% \begin{enumerate}
%     \item comparison with on-robot ToF sensors \cite{kumar_speed_2019}
%     \item human shape aware detection enables circumventing self-collision checking %don't have to self filter like in the tof rings in the shitij protocol
%     \item processing in lidar image space is faster than raw point cloud processing
%     \item can leverage state of the art deep learning based approached on lidar images
%     \item capture data from diverse body shapes and clothing material in an industrial environment
%     \item analyse the viability of a vertical and horizontal fov lidar for safety based applications
% \end{enumerate}

%Relation to our previous work:
%Used tof rings in the past, researching a newer sensing modality within an on-robot sensor placement setup. 

\section{Methodology}
This section covers the various components involved in the experimental process. The goal of of the setup was to explore the usage scenarios for 3D lidars such as the Ouster OS-0-32 in an industrial shop floor environment. This environment was comprised of mostly static objects (workbenches etc.) with a limited number of dynamic objects (humans \& robots) within the lidar FoV. 

\begin{figure}[h]
\includegraphics[width=\linewidth]{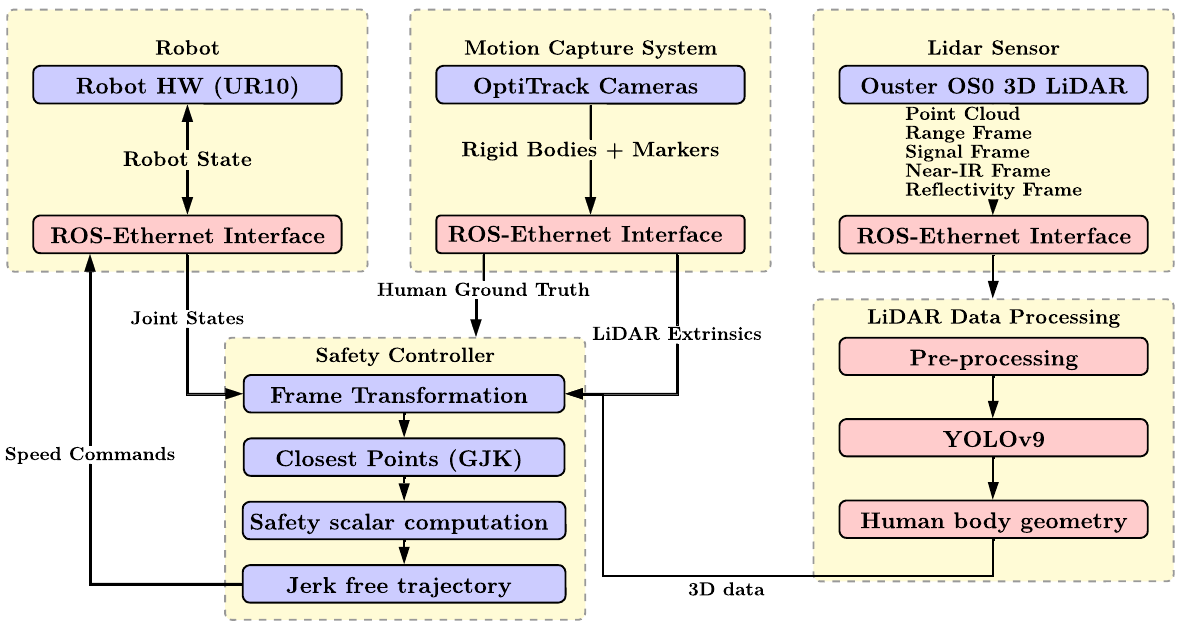}
\caption{Control schema showing the complete system, our communication is powered by Robot Operating System (ROS).}
\end{figure}

\subsection{Setup \& Calibration}
In this step, the focus was on achieving a time synchronization between the heterogeneous data streams emanating from the lidar sensor, motion capture, and the robot control box. Synchronization relied on a local high speed Ethernet based network that exhibited an average latency of approximately 0.25 milliseconds (round-trip time). Therefore, it was assumed that the delta time between the time of arrival of data packets and the time of origin was negligible. Finally, an asynchronous time synchronization on the various streams was performed. The inter-stream time delta of the synchronization was 5 milliseconds. In the calibration procedure, the main goal was to obtain the rigid body transformations of all the sensing data in a common reference frame. Customized rigid body marker-sets we developed for the motion capture system. These marker were affixed to the lidar, pedestal of the robot, and also on the skeleton tracking body suit worn by a human participant. This step provided a coarse calibration, however, for a better estimate of the lidar extrinsics an optimization based point-set alignment was used as shown in \cite{umeyama_least-squares_1991}.

\subsection{Data Collection and Labeling}
Once a synchronized and calibrated setup was achieved, the incoming data was recorded over the network on a local disk storage. The following data fields were focus on:
\begin{itemize}
    \item \textbf{Lidar Data} at 20 Hz: Point-cloud, Refelectivity, Signal, Near-Infrared and Stacked images
    \item \textbf{Motion Capture Data} at 120 Hz: Rigid-bodies and 3D marker locations
    \item \textbf{Robot State Data} at 125 Hz : Joint positions and velocities
\end{itemize}

\begin{figure}[H]
    \centering
    \includegraphics[width=8.2cm]{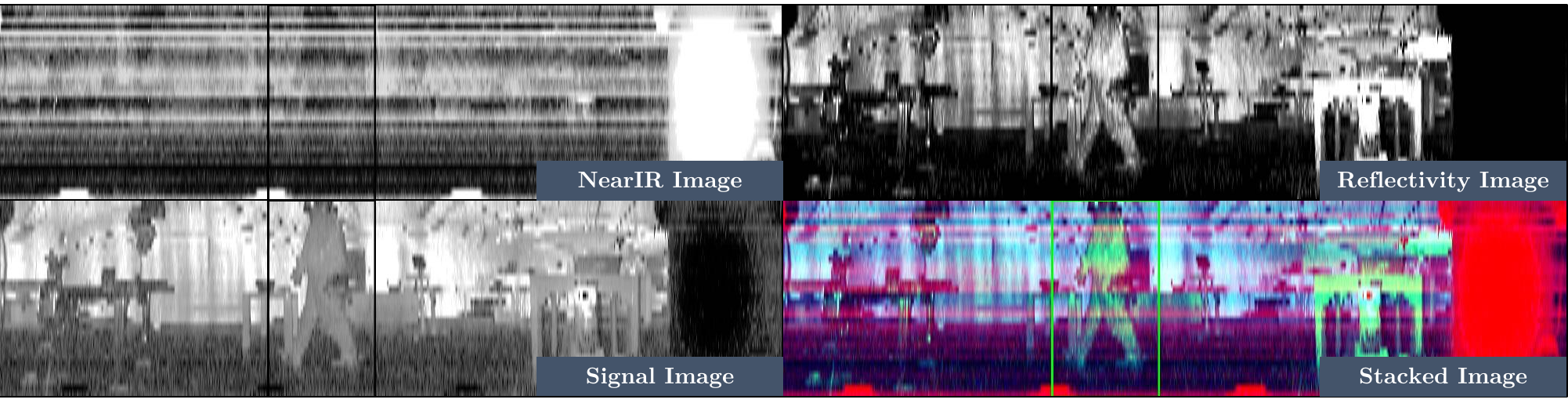}
    \caption{Clean samples of the reflectivity, signal, near-IR, and a depth-wise stacked image of the first three. The annotation is overlayed on the grayscale images in black and in green on colored images.}
    \label{fig:enter-label}
\end{figure}

After the data was collected, it was then processed for labeling and downstream tasks such as low-level image processing and classification. The bulk of process, comprised of pre-processing steps applied to the lidar point-cloud, and image quadruplet obtained from the lidar data. First, the lidar point-cloud images were ``\textit{destaggered}" as mentioned in the documentation provided by manufacturer in \cite{ouster_ouster_2022}. The main idea behind this step was to remove the time offset from each element of the lidar data (point-cloud and images). Afterwards, the image quadruplet was subjected to bit depth down-sampling from 16-bit to 8-bit image data. As the image resolution was $1024\times32$, the images had to be resized to $1024\times256$ by applying bi-linear interpolation. The images were then subjected to auto-exposure adjustment and histogram equalization as part of pre-processing. The images were then annotated in a semi-automated fashion with bounding boxes in MSCOCO format \cite{lin_microsoft_2015}. For semi-automation, the static nature of the environment was exploited to remove a large number of points by background removal and applying statistical outlier rejection on the remaining points. Afterwards, noisy bounding-box labels were generated by re-projecting the non-stationary 3D points into the images. Ultimately, the bounding boxes were hand tuned.

\subsection{Network Training and Inference}
The YOLOv9 \cite{wang_yolov9_2024} object detection network was selected to annotate bounding boxes around the human body shape in the lidar data (to image quadruplet only). For this step, two datasets were developed from the data collected during multiple experiments. The two variants comprised of single-channel annotated reflectivity images and multi-channel annotated images where reflectivity, signal, and near-infrared images were stacked depth-wise forming a tensor. It must be noted, that the images in the datasets represented only a subset of the total data recorded during the experiments. A larger full version of the dataset will be made available for the research community. Transfer learning was performed on a pre-trained variant of YOLOv9 called ``\textit{YOLOv9-C}" that had fewer parameters than the largest YOLOv9 variant called ``\textit{YOLOv9-E}". The network was selected due to it's state-of-the-art performance and efficiency as shown in \cite{wang_yolov9_2024}. Both datasets consisted of 14,000 annotated images, the dataset split was selected as 80\% training and 20\% validation. For testing, new dual variants of single-channel and multi-channel datasets (comprised of unseen data by the network) were prepared. As safety is one of the key challenges in pHRI \cite{villani_survey_2018}, it is important to analyze every labeled and unlabeled image by the network at inference time to determine its suitability in a high stakes scenarios. Therefore, the test-set created was representative of one full trial performed by a human subject during the experiment, and validated using the fine-tuned network. For training, the stochastic gradient descent (SGD) optimizer with a momentum of 0.937 was selected. The batch size and epochs were chosen to be 16 and 50, respectively.
%TODO - Missing training parameters??????????????????????
% \begin{table}[H]
%     \centering
%     \begin{tabular}{|c|c|}
%          \textbf{Parameter} & \textbf{Value} \\
%          Epochs & 50 \\
%          Batch size & 16 \\
%          Optimizer & SGD \\
%          Momentum & 0.937 \\
%          Regularization Term & 0.0005
%     \end{tabular}
%     \caption{Training Hyper-parameters}
%     \label{tab:hyper}
% \end{table}
% \vspace{-20pt}
\subsection{Human point cloud extraction}
In this phase of the pipeline, the annotations provided by the aforementioned network at inference time were used. As the spatial structure of the lidar frame (comprising of point-cloud and image quadruplets) allowed for a bi-directional mapping between the images and the point-cloud. The bounding-box rectangles were projected into a corresponding point-cloud and points external to the region of interest were pruned. This reduced the total number of points from $1024\times32$ to approximately $20\times50$ (based on the largest possible size of the bounding box). Then, plane-segmentation and DBSCAN \cite{ester_density-based_1996} clustering were used to extract the points associated with the human body shape.  

\begin{figure}[H]
\centering
\includegraphics[width=4cm]{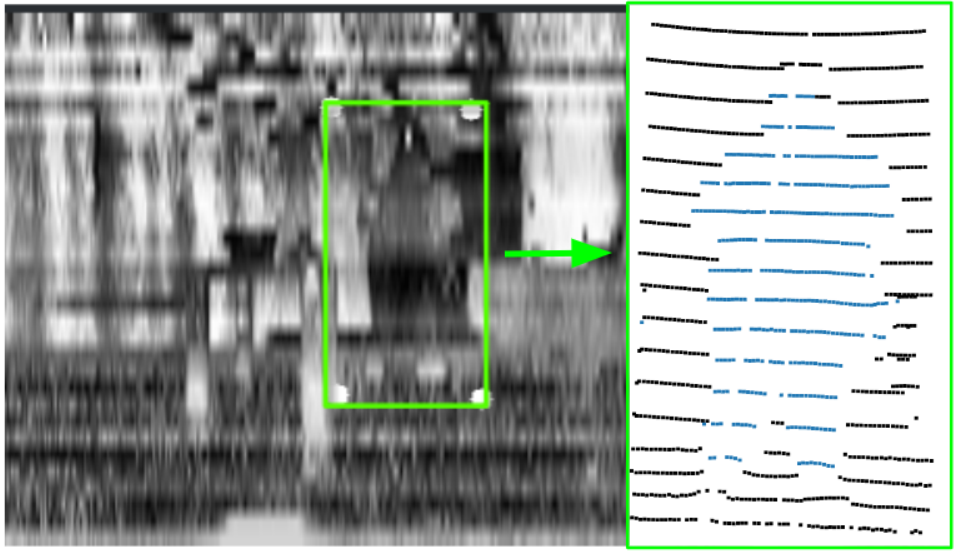}
\caption{An image showing human shape geometry extraction using a point-cloud along with a bounding determined from a reflectivity image. Points in blue represent the human body, points in black are rejected as background.}
\end{figure}

\subsection{Speed and Separation Monitoring Algorithm}
The method towards the implementation of the Speed and Separation Monitoring (SSM) was derived from\cite{kumar_speed_2019}. This work defines the 3D geometrical (in a common reference frame as the robot) representation of the human operator in the robot's workspace. A scene graph was constructed and closest pair of point queries between the human and the robot were performed. The algorithm of choice for such tasks was the ``GJK" algorithm \cite{gilbert_fast_1988} which is widely used for such applications. The closest pair of points allowed for the computation of the minimum distance vector. This vector was used to compute the protective safety distance (threshold or a barrier around the robot) to trigger the robot stop behavior. As stated in \cite{iso_isots_2022} and \cite{kumar_speed_2019}, the speed and separation monitoring equation is given by:

\begin{equation}\label{eq:1}
S_{safety}(t_{0}) = V_{human}\cdot(t_{r} + t_{s}) + V_{robot}\cdot t_{r} + C + Z_{s} + Z_{r}
\end{equation}

$Z_{s}$ \& $Z_{r}$ represent the position measurement uncertainties for human and robot respectively. These values were obtained from the data-sheets of the robot and the exteroceptive sensing equipment used. $C$ is the intrusion distance which is defined by \cite{iso_iso_2010}, in essence it represents the threshold at which an obstacle is successfully detected. $t_r$ and $t_s$ represent the control loop processing time and the time required by the robot to come to a full stop, respectively. These time values were also be obtained from the robot's (UR10) data-sheet or can be empirically estimated. The robot stopping time aka $t_s$ can also be tuned but should be lower bound by the worst case stopping time. To achieve jerk free stop behavior, the online trajectory generation library \cite{berscheid_jerk-limited_2021} was used.

As governed by the standard \cite{iso_isots_2022}, most terms in the equation can be substituted with constants, the only quantity which is non-trivial to compute is $V_{robot}$. Thus, a fairly simple but general algorithm is proposed to compute the directed velocity of the robot towards the human based on velocity kinematics of the robot:

\begin{algorithm}
\caption{Algorithm to compute directed $\textbf{V}_{robot}$}\label{alg:three}
\KwData{Closest points: $P_{human}$, $P_{robot}$\\ 
Workspace limit: $W_{max}$\\ 
Routine: ``$jacobian( )$" to compute robot jacobian}
\KwResult{$V_{robot}$ in direction of the operator}
 $\vec{S} \leftarrow P_{human} - P_{robot}$\; $\vec{z} = (0,0,1)^T$\\
  \eIf{is\_valid($\vec{S}$)}{
   $\vec{f}$ = $normalize(\vec{S})$\;
   $\vec{r}$ = $\vec{z}\times\vec{f}$\\
   $\vec{u}$ = $\vec{f}\times\vec{r}$\\
   $^{w}\textbf{T}_{r}$ =
   $\begin{psmallmatrix}
   \begin{smallmatrix}\vec{\textbf{r}}\end{smallmatrix} & \begin{smallmatrix}\vec{\textbf{u}}\end{smallmatrix} & \begin{smallmatrix}\vec{\textbf{f}}\end{smallmatrix} & \begin{smallmatrix}\textbf{P}_{robot}\end{smallmatrix}\\
   0 & 0 & 0 & 1
   \end{psmallmatrix}${\tiny\textit{directed transform along z in world coordinates}}\\
   \vspace{2pt}
   $\textbf{J}_{robot} = jacobian(\textbf{q}, ^{w}\textbf{T}_{r})$ \\
$\begin{bmatrix}^{w}\textbf{v}_{twist}\end{bmatrix}_{6\times1}=\textbf{J}_{robot}\cdot \dot{\textbf{q}}$ \\
   $^{w}\textbf{T}_{h} = ^{w}\textbf{T}_{r} \cdot$ $\begin{psmallmatrix}1&0&0&0\\0&1&0&0\\0&0&1&\lVert\vec{\textbf{S}}\rVert\\0&0&0&1\end{psmallmatrix}$\\
   \vspace{2pt}
   $^{h}\textbf{v}_{r} = {^{w}\textbf{T}_{h}}^{-1} \cdot \begin{bmatrix}^{w}\textbf{v}_{twist}\end{bmatrix}_{4\times4}$\\
   % $\textbf{V}_{robot} = {^{h}\textbf{v}_{r}}\langle \textbf{z} \rangle$\\
   ${V}_{robot} = sign({^{h}\textbf{v}_{r}}\langle \textbf{z} \rangle)
   \cdot
   \lVert
   {^{h}\textbf{v}_{r}}\langle \textbf{z} \rangle
   \rVert$\\
   $return$ ${V}_{robot}$ 
   }{$return$ $-1$
   }
\end{algorithm}

Using the signed scalar given by $V_{robot}$, it can be substituted in \eqref{eq:1} to obtain the safety distance. The safety distance $S_{safety}$ and the magnitude of minimum distance vector $\lVert\vec{S}\rVert$ can be used to compute the speed scaling commands sent to the robot controller by the following computation: 

\begin{equation} \label{eq:2}
    \rho_{scaling} = \frac{max(\lVert\vec{S}\rVert-S_{safety}, 0)}{W_{max}}
\end{equation}

 such that $\rho_{scaling} \in [0,1]$ and $W_{max}$ is the distance limit in meters for the robot workspace beyond which sensor readings are clamped. $\rho_{scaling}$ represents the scaling factor to control the speed of the robot. This factor can be used to uniformly scale the joint velocities $\dot{\textbf{q}}$ of the robot as shown in \cite{kumar_speed_2019}.

\section{Experiment}

\begin{figure}[H]
\centering
\includegraphics[width=6.5cm]{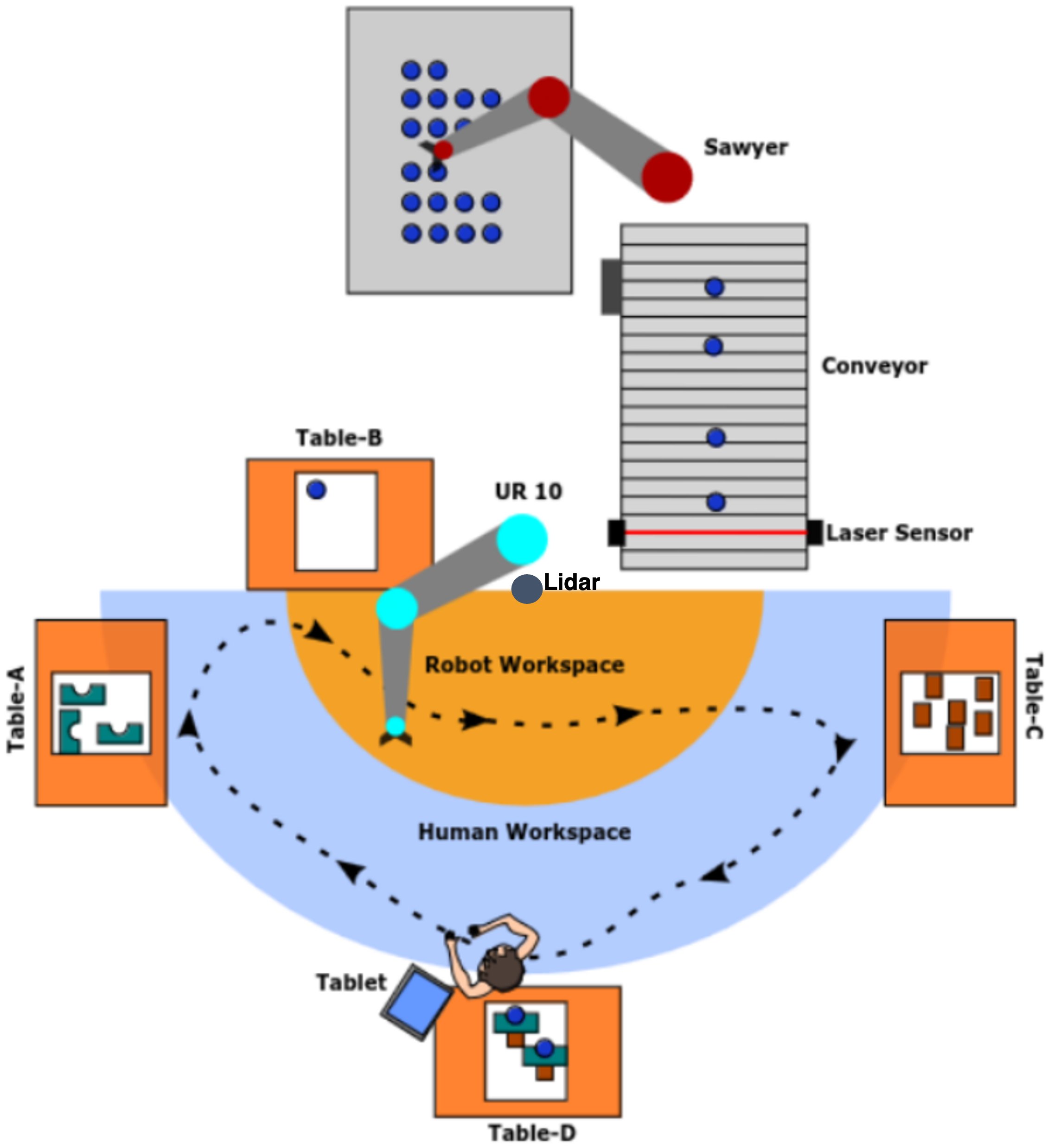}
\caption{Flow diagram of the experiment performed as prescribed by the authors in \cite{savur_physiological_2022}
}
\end{figure}

The experimental configuration comprised pf distinct tasks allocated to both the human and the robot, emulating an assembly line scenario. Specifically, a collaborative robot was responsible for extracting a component from a pallet and depositing it onto a conveyor belt. Subsequently, a UR-10 robot retrieved the necessary component from the conveyor and situated it within a bin accessible to the human operator, both the human and UR-10 coexisted within the same workspace. There were four stations in the shared workspace. The human worker walked to each of these stations completing the assembly of a PVC coupling. One of these parts was provided by UR-10. Once the assembly of a single part was completed, the human was required to deposit the item at a designated location and perform an arbitrary task at the same station. This process was repeated 24 times in one trial. A lidar captured point clouds and multi-channel imagery throughout the experiment. Each worker participated in 6 trials, in 3 of those trials, they were required to wear a high visibility jacket. Additionally, the entire workspace was monitored with a motion capture system which possessed 13 cameras that flooded the workspace with 940 nm infrared light.

\section{Results and Discussion}

\subsection{Dataset}
\begin{figure}[H]
    \centering
    \includegraphics[width=8.5cm]{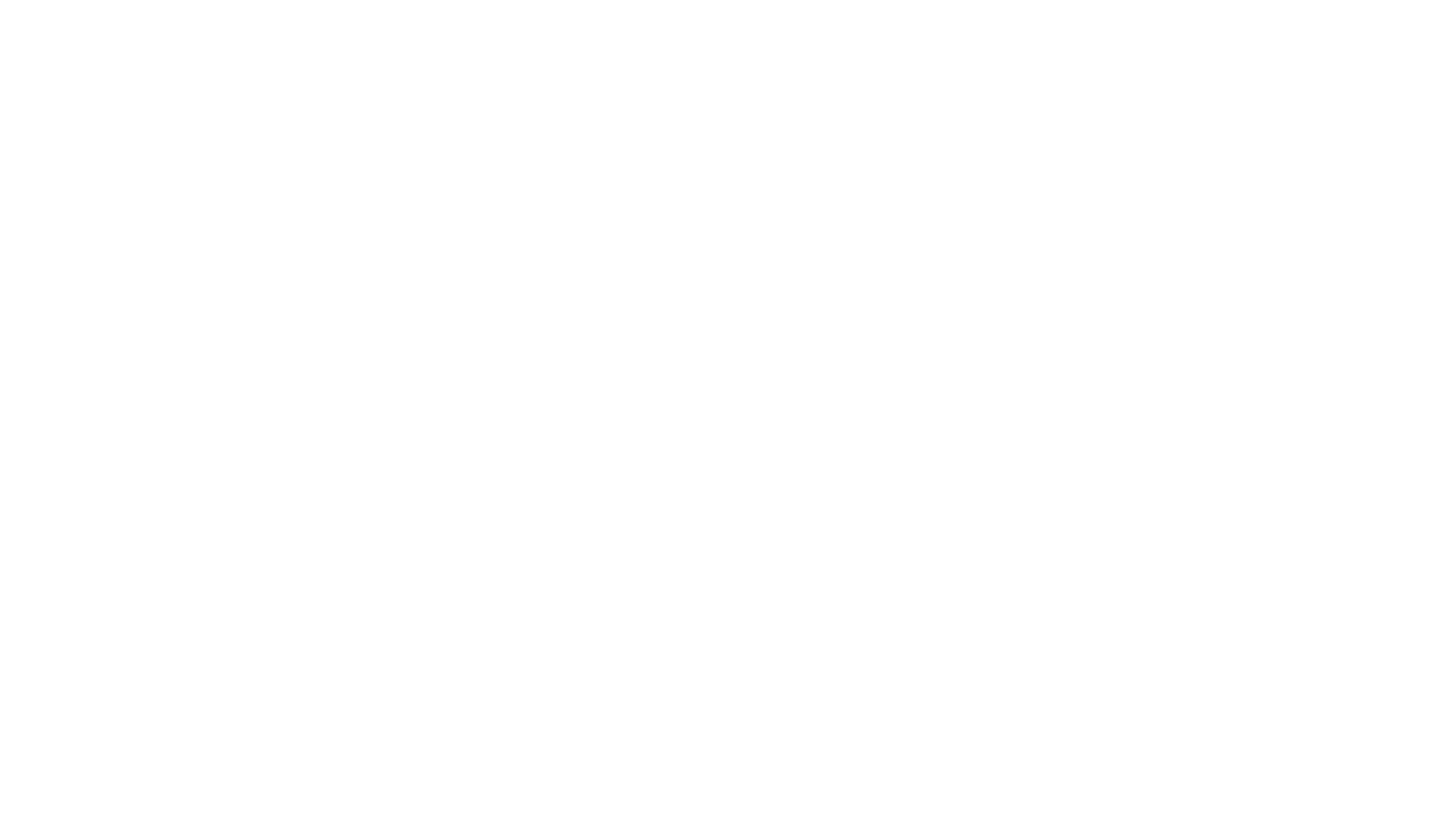}
    \caption{A plot and word-cloud showing the participant attribute distribution. Left: the weight-height and age distribution of the participants. Right: a word cloud showing a distribution of clothing types, materials and colors.}
    \label{fig:participants}
\end{figure}

The experiment involved 17 participants, 29\% of the participants were of female sex and the remaining 71\% were of male sex. The variety of clothing, fabrics, and colors worn by participants were recorded. A world-cloud image to represent this diversity is illustrated in Figure \ref{fig:participants}. The most common clothing color, fabric, and type were black, cotton and jeans with hoodie, respectively. This gave a minor insight that clothing worn by operators in shop-floors could also be dark in color and of cotton material. \textit{Hence, a lidar should be able to detect these materials based on any arbitrary surface reflectivity exhibited by them.}
The weight of the participants ranged from 49 kg to 136 kg, and the height ranged from 1.5 m to 1.85 m. It is vital for any learning based model to be aware of varying body geometries in a shop-floor environment. The approval for Human subject research was granted by Rochester Institute of Technology. (Approval number: \textbf{21081267})

After the data collection, pre-processing was applied and the training and validation sets (for ``single-channel" and ``multi-channel") were prepared with about 12,000 and 2200 images, respectively. 

\begin{figure}[H]
    \centering
    \includegraphics[width=4cm]{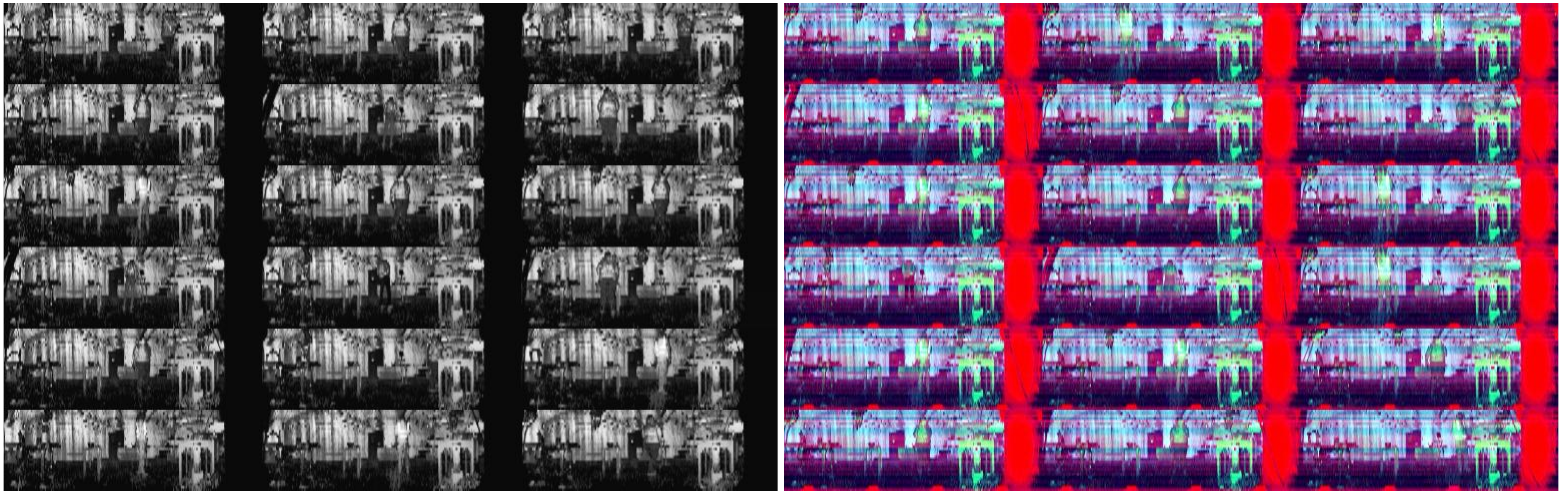}
    \caption{Tiled layout of 18 samples randomly drawn from each training dataset. Left: a snapshot of the single-channeled dataset built with reflectivity images. Right: a snapshot of the multi-channeled dataset built with depth-wise concatenation of reflectivity, signal, and near-infrared images.}
    \label{fig:datasets}
\end{figure}

\subsection{Quantitative Results}
\begin{figure}[H]
    \centering
    \includegraphics[width=8.5cm]{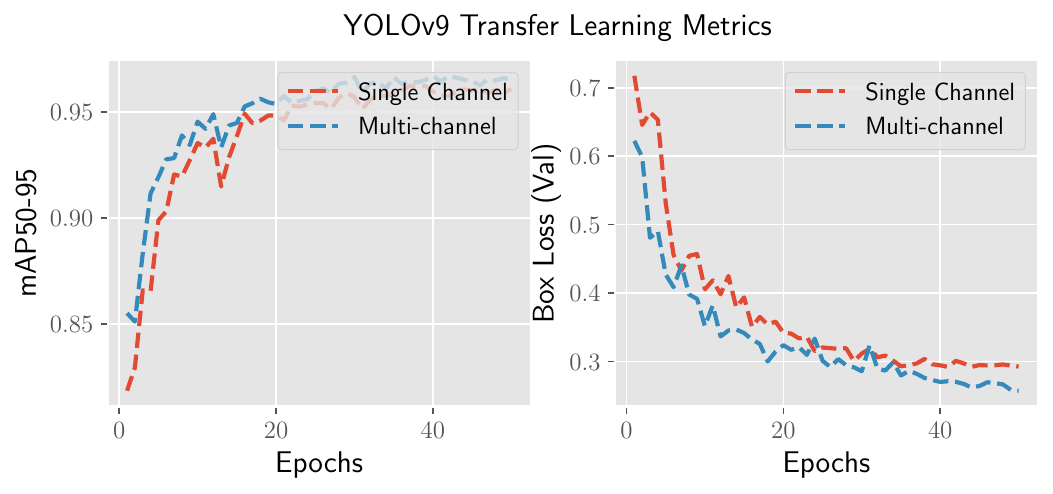}
    \caption{Plots showing the validation metrics during YOLOv9 fine-tuning on the datasets prepared namely Single-channel \& Multi-channel datasets.}
    \label{fig:yolo-training}
\end{figure}

The two previously mentioned datasets were used to train the YOLOv9 object detector in a binary detection mode. The validation curves during training session are shown in Figure \ref{fig:yolo-training}. During the ``multi-channel" training it was observed that the YOLOv9 network converged faster and exhibited a higher mAP50-95 validation score. 

\begin{figure}[H]
\centering
\includegraphics[width=7cm]{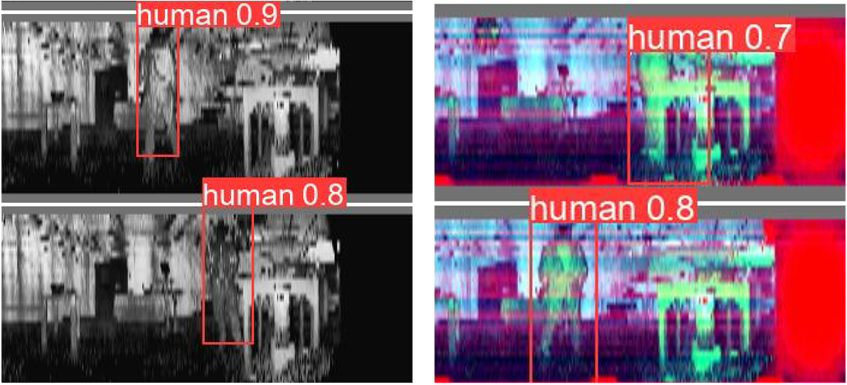}\\
\includegraphics[width=8.5cm]{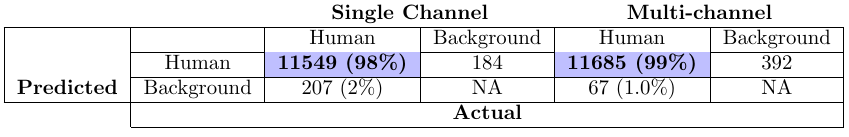}\\
\noindent
\includegraphics[width=4.3cm]{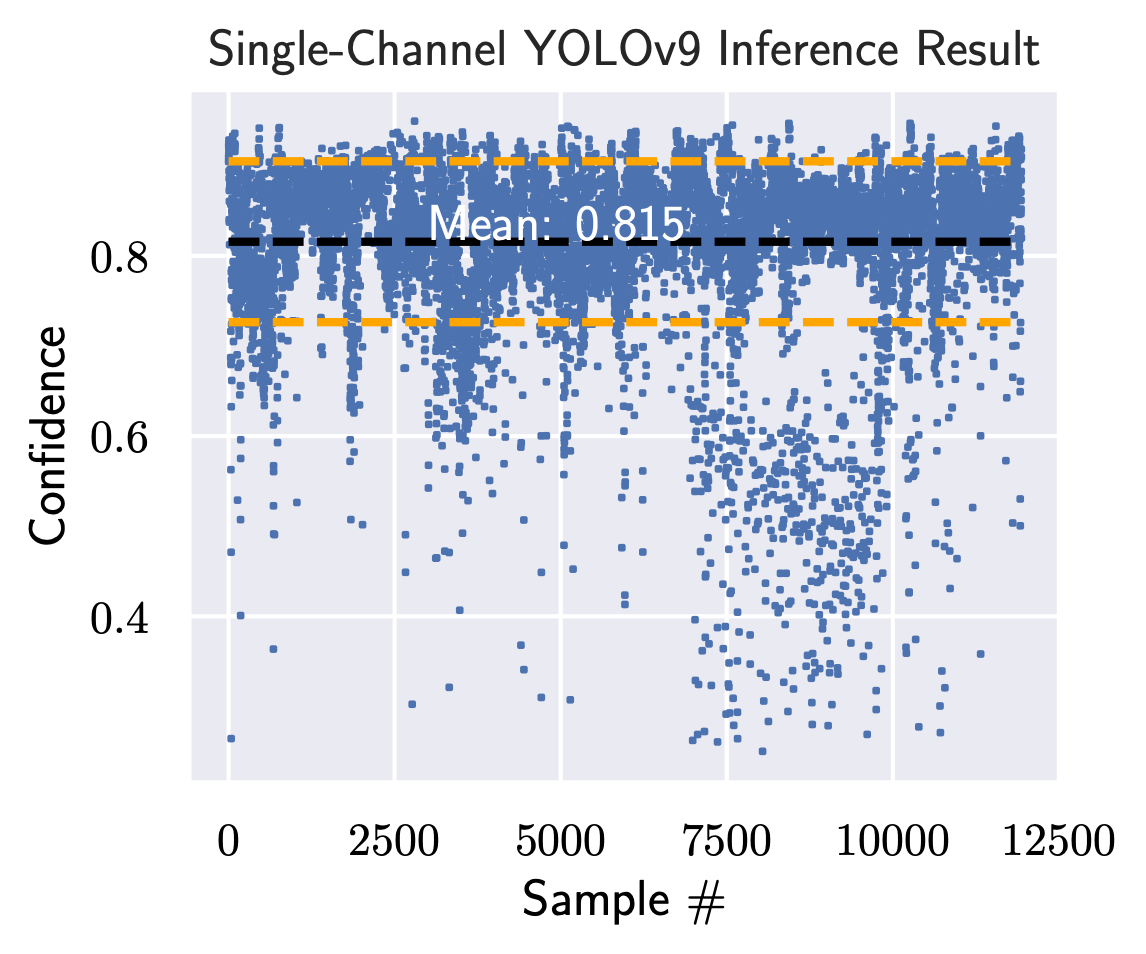}
\hspace{-0.6cm}
\includegraphics[width=4.3cm]{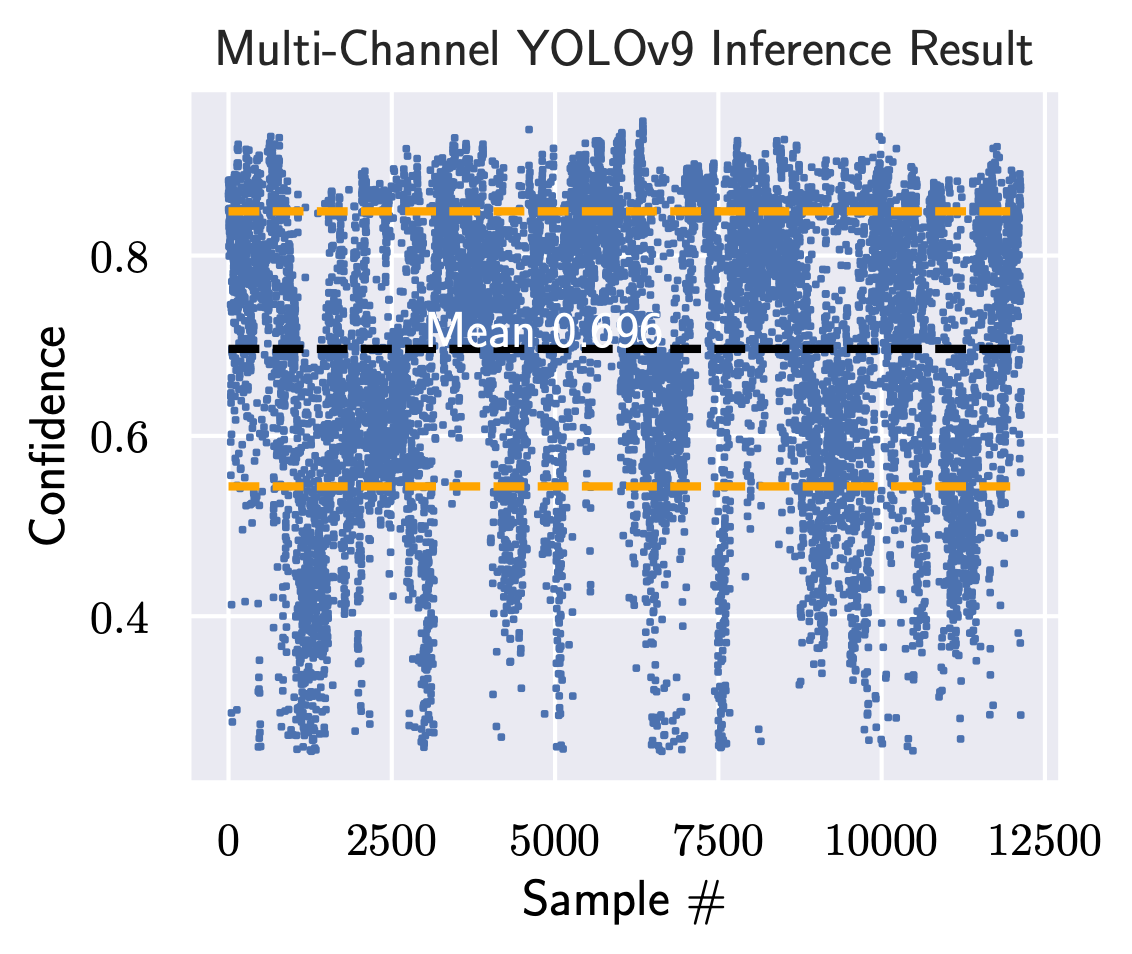}
\caption{Figures showing inference examples, confusion matrix, and confidence scatter plots on the test sets.}
\label{fig:yolo-testing}
\end{figure}

After training the network, inference was performed on unseen lidar sequences of 12,500 samples for both variants. In figure \ref{fig:yolo-testing}, it was observed that the ``multi-channel" variant performed approximately 1\% better than the ``single-channel" variant. However, it was noted that the classifier confidence during inference was more robust during ``single-channel" inference. On analyzing the spread of the confidence values of the classifiers, it was found that the multi-channel detector was measurably less certain than the single-channel variant.
% \begin{table}[H]
%     \centering
%     \includegraphics[width=8.5cm]{confusion_matrix_case2024-cropped.pdf}
%     \caption{Confusion matrix for single-channel and multi-channel variants}
%     \label{tab:confusion_mat}
% \end{table}
To measure the accuracy of the lidar for pHRI scenarios, the closest pair of points between the human operator and the robot were recorded with the lidar and the motion capture system simultaneously. For the on-robot base mounted 3D lidar, the root mean square error (RMSE) was more than 4 times lower than on-robot time-of-flight sensing rings in \cite{kumar_speed_2019}. The margin of error was found to be lower bounded by 3mm as reported by the manufacturer.
\begin{table}[H]
\centering
\begin{tabular}{c|c|c|c|}
\cline{2-4}
    & & Lidar (ours) & ToF Rings \cite{kumar_speed_2019} \\ \cline{2-4}
    & \textbf{RMSE (m)} & \cellcolor{blue!25}{\textbf{0.0605}} & 0.25 \\ \cline{2-4}
\end{tabular} 
\label{tab:mindist}
\caption{RMSE comparison for lidar and time-of-flight sensing rings}
\end{table}
\begin{figure}[H]
\includegraphics[width=\textwidth/2]{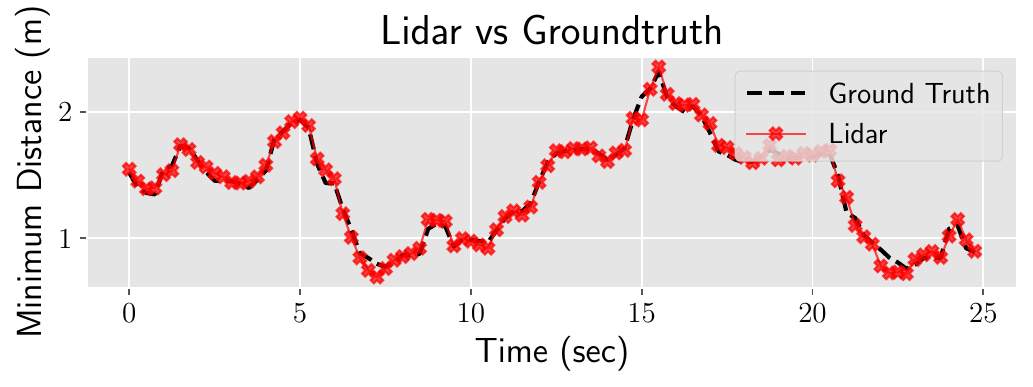}
\caption{Plot showing the minimum-distance comparison between data acquired from the lidar and motion capture system overlay-ed on top of each other.}
\label{fig:min-dist-plot}
\end{figure}

\begin{figure*}[h]
\includegraphics[width=\textwidth]{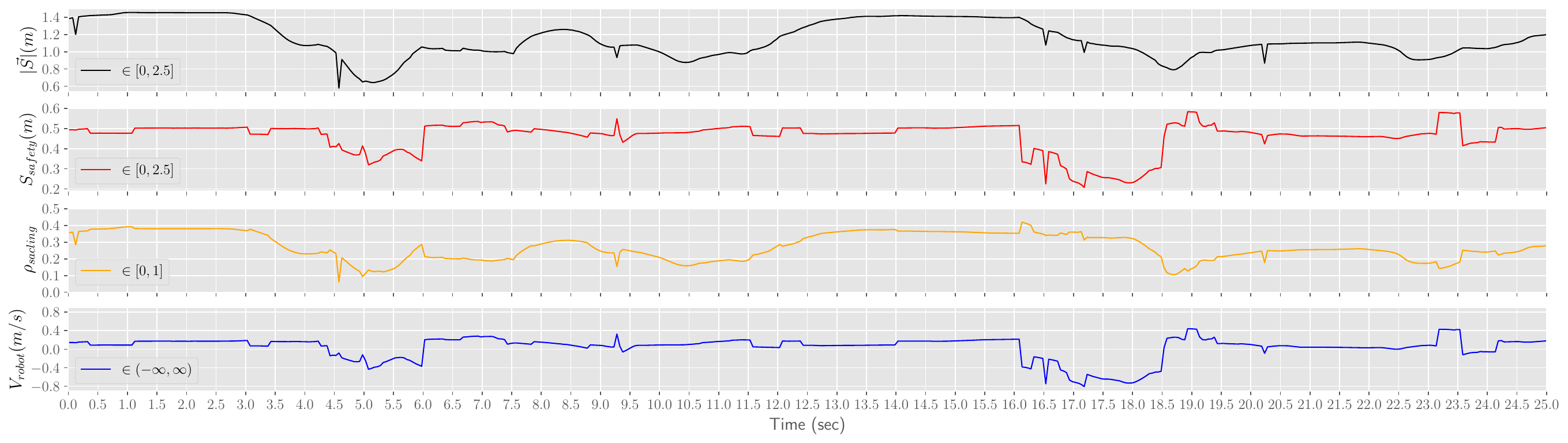}

\caption{Time series (25 seconds) plots showing the results of the directed robot velocity computation (bottom most in blue) towards the human along with the minimum distance (top in black), safety distance threshold (second from top in red) computed with the SSM Equation \cite{iso_isots_2022} and the speed scaling factor (third from top in yellow) used for modulating the operational speed of the robot.}
\label{fig:ssm-results}
\end{figure*}

The results from the safety algorithm are presented in figure \ref{fig:ssm-results}. A 25 $second(s)$ long recording of the results was analyzed; the directed robot velocity \textbf{$V_{robot}$} was found to be proportionally tracked by safety distance $S_{safety}$ due to its linear dependence on the prior. $V_{human}$ was set to $1.6m/s$ (prescribed by \cite{iso_isots_2022}) and the remaining terms in equation \ref{eq:1} were construed from the robot's datasheet. Between 18.5 and 19.0 seconds marks, the speed scaling term $\rho_{scaling}$ decayed when the minimum distance $\lVert\vec{S}\rVert$ violated or tended towards $S_{safety}$. It should also be noted that $\rho_{scaling}$ was always below $0.5$, as the human subject was always within 1.5 meters ($<W_{max}$) of the robot. Furthermore, even though there was no smoothing and filtering applied to the data, $\lVert\vec{S}\rVert$ computed from lidar data was significantly smoother than in \cite{kumar_speed_2019}, where an exponential filter was used. \textit{As closest distance between two articulated bodies can tend to vary at higher frequencies, the capability of the system to provide a relatively smoother metric virtually eliminates the need of low-pass filters that can introduce time lag in the controller. This is vital in high stakes scenarios where safety is the main goal.} 

% \begin{figure}[H]
% \includegraphics[width=4cm]{rolling-timestamp.pdf}
% \includegraphics[width=4cm]{ssim.pdf}
% \end{figure}

% \begin{figure}[H]
% \centering
% \end{figure}

	% 	Lidar vs Optitrack: rmse from this paper
  
	% Image similarity:
	% 	Image frame similarity matrix
	% Hypothesis:	images are spatially correlated but have photo-metric variations, therefore can be used as multi-channels in a network due to non-redundant information
	% Rolling shutter results for the lidar
	% 	sample capture timestamp monotonic graph
 %  Figure x shows the monotonic nature of the lidar recordings. This can cause a rolling shutter effect. While the lidar records at relatively good frequency, it may be susceptible to creating artifacts if an object in the frame moves fast enough from one point to another before the rolling shutter has scanned the entire frame. This may create a situation which will make the moving object appear like it has teleported. Swiftly moving objects can also appear distorted as they may have been recorded at staggered intervals.
  
	% Min-dist comparison
	% 	Relates to staggering/de-staggering 

\subsection{Qualitative Results}
\begin{figure}[H]
\includegraphics[width=8.5cm]{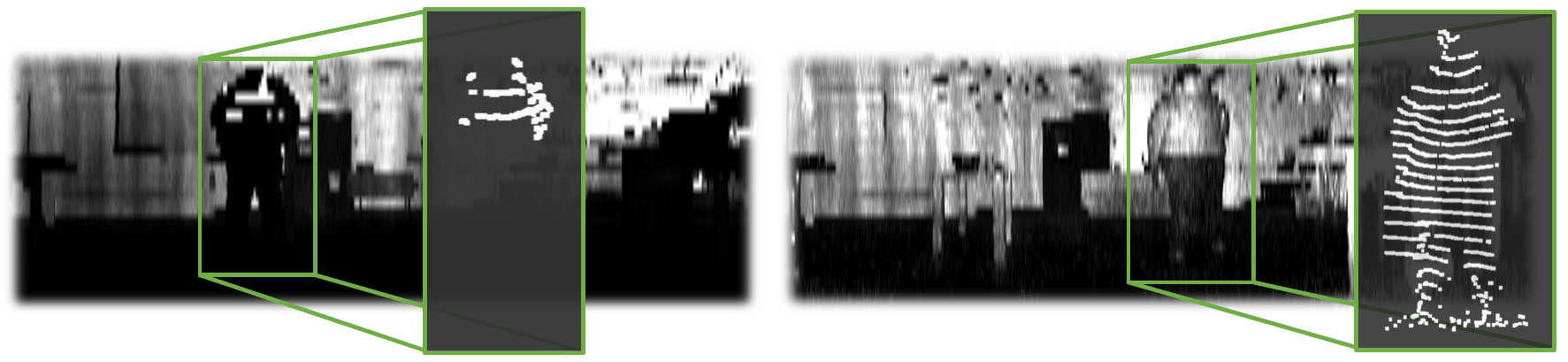}
\caption{On left, lidar reflectivity image with holes with its corresponding 3D point-cloud in perspective view. On right, a healthy sample of the reflectivity image with its 3D point cloud.}
\label{fig:holes}
\end{figure}

During data collection, it was observed that the response of the lidar was poorer in certain scenarios where the participants were wearing significantly darker clothes even in close proximity to the robot. It was found that the reflectivity and range images provided by the lidar exhibited the presence of holes. As a consequence, the point cloud lacked the 3D information associated with the human's shape geometry (points were missing from the point cloud). This phenomenon is illustrated in a side by side comparison shown in figure \ref{fig:holes}. It should be noted that in the left half of the figure, the participant was wearing a high-reflectivity vest with black cotton garments underneath. Only the points associated with the reflective vest were reported by the lidar. 

\begin{figure}[H]
    \centering
    \includegraphics[width=6cm]{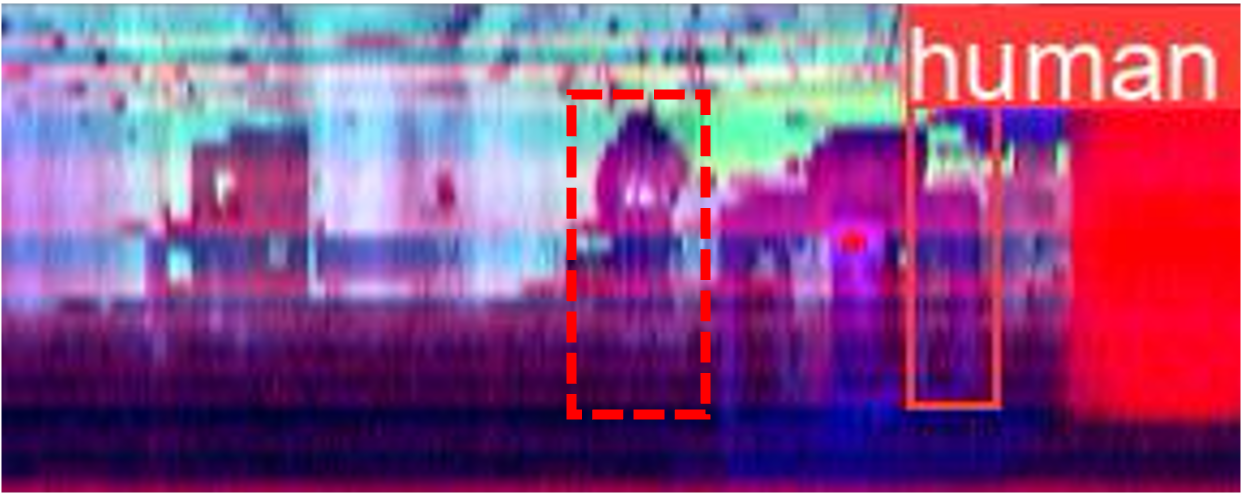}
    \caption{A miss-classification performed by the ``multi-channel" variant due to room layout change and ambiguous texture patches in the input image. The correct bounding box is drawn in a dashed bounding box.}
    \label{fig:failure}
\end{figure}

Another limitation was observed, wherein the ``multi-channel" variant performed poorly after the floor layout changed. This limitation is shown in figure \ref{fig:failure}, this can be explained due to a distribution shift, the network is biased directly on the metrics associated with the photons scattered in the environment. The colored patches in the image can also create ambiguous textures that can confuse the network. It should be noted, that in this image the participant is wearing a reflective vest. This can be also be addressed by a higher resolution lidar such as OS-0-128 where the base resolution is $1024\times128$, hence the effect of up-scaling will not create aliasing artifacts.

\begin{figure}[H]
\centering
\includegraphics[width=4cm]{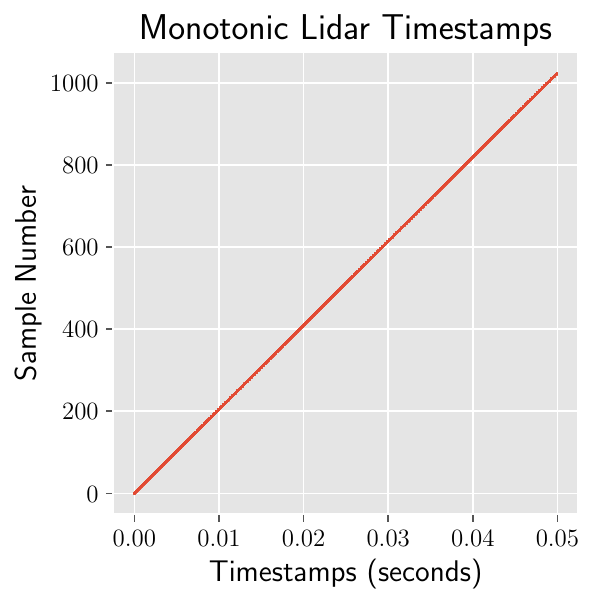}
\includegraphics[width=4cm]{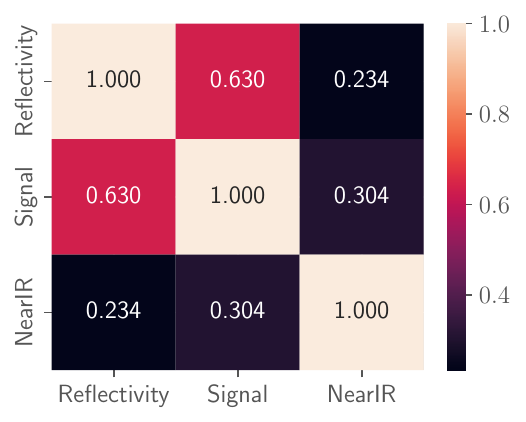}
\caption{Left, the timestamps of the sequential readout perform by the lidar. Right, a matrix showing the relative structured similarity index metric (SSIM) of the lidar images.}
\label{fig:lidar-prop}
\end{figure}

Figure \ref{fig:lidar-prop} shows the monotonic nature of the lidar recordings on the left. This can cause a rolling shutter effect; while the lidar records at relatively high frequency (at 20 $Hz$), it may be susceptible to creating artifacts if an object in the frame moves fast enough from one point to another before the rolling shutter has scanned the entire frame. This may create a situation where the moving object appears to have teleported in the recorded frame. Swiftly moving objects can also appear distorted as they may have been recorded at staggered intervals. To explain for a slightly higher validation and inference results by the ``multi-channel" variant, the structured similarity index metric (SSIM) matrix was used. As shown on the right in \ref{fig:lidar-prop}, as the relative SSIM of each image type is significantly below 1.0, there is a likelihood that the network can extract additional features from the added channels. If the features were redundant, the relative SSIM (the matrix elements would diffuse more) would be closer to 1.0. However, the Near-IR channel tends to exhibit higher ambient noise than other images, but can be useful in close proximity scenarios where objects are closer than the minimum measurable distance by the lidar.  

% multichannel results
% Single channel results
% Lidar image with holes
% Talk about participants with dark clothes and image holes 

\section{Conclusions \& Future Work}
The on-robot base mounted lidar can significantly outperform on-robot time-of-flight sensing rings due to the 3D point-cloud and 2D image data. Furthermore, the bi-directional $2D\Leftrightarrow3D$ mapping enables for higher level tasks such as object detection on images and subsequently, region-of-interest extraction on corresponding point-clouds. This leads to a more efficient perception pipeline as image based backend(s) can be used to bootstrap detection networks while pruning the 3D search space. Also, due to this capability, we were able to semi-automate bounding box annotation for our datasets. In future, this can enable the application of techniques such as continual learning \cite{wang_comprehensive_2024}.

The lidar also exhibits some limitations due to the presence of holes in the image channels which affect the quality of the point-cloud. Therefore, in a shop-floor it is vital to wear high-reflective markers such as vest and helmets as they can alleviate the presence of holes in the lidar data. Ultimately, we can conclude that the use of the 3D lidar in close proximity pHRI scenarios is viable, as long as steps are taken to prevent sensing failures and pitfalls.\\
For future works, the first step is be to develop a target network that can directly handle the input sizes provided by the lidar and is designed to work with 16-bit precision. To overcome the distribution shift problem, the channel order can be randomized while also introducing small changes in the floor layout so that the network becomes more robust. The changes required would be quite small, as a shop-floor environment is more static than an outdoor scenario. Exploring deep learning based image up-scaling techniques such as \cite{lim_enhanced_2017} and usage of more advanced sensing hardware (OS-0-128) will also provide us with more reliable inference. Another downstream task that we are already working is instance segmentation, we are currently working on developing mask annotation for the lidar images.
For the safety controller, leveraging directed velocity of the human operator towards the robot will also aid the safety barrier to be relaxed in situations where the human is moving away from the robot. As we assume $V_{human}$ to be a positive constant, it implies that operator is always moving in the direction of the robot with a constant velocity. Therefore, measuring the velocity of the operator in real-time will be beneficial for robot productivity without sacrificing operator safety. 

%%%%%%%%%%%%%%%%%%%%%%%%%%%%%%%%%%%%%%%%%%%%%%%%%%%%%%%%%%%%%%%%%%%%%%%%%%%%%%%%

%%%%%%%%%%%%%%%%%%%%%%%%%%%%%%%%%%%%%%%%%%%%%%%%%%%%%%%%%%%%%%%%%%%%%%%%%%%%%%%%


% % Appendixes should appear before the acknowledgment.

% % % \section*{ACKNOWLEDGMENT}

% % % % The preferred spelling of the word ÒacknowledgmentÓ in America is without an ÒeÓ after the ÒgÓ. Avoid the stilted expression, ÒOne of us (R. B. G.) thanks . . .Ó  Instead, try ÒR. B. G. thanksÓ. Put sponsor acknowledgments in the unnumbered footnote on the first page.

%%%%%%%%%%%%%%%%%%%%%%%%%%%%%%%%%%%%%%%%%%%%%%%%%%%%%%%%%%%%%%%%%%%%%%%%%%%%%%%%

%References are important to the reader; therefore, each citation must be complete and correct. If at all possible, references should be commonly available publications.

\bibliography{CASE_2024_references}
\bibliographystyle{plain}

\end{document}